\documentclass[11pt,a4paper]{article}
\usepackage[hyperref]{emnlp2020}
\usepackage{times}
\usepackage{latexsym}

\usepackage{booktabs}
\usepackage{bm}
\usepackage{bbm}
\usepackage{amsmath}
\usepackage{appendix}

\usepackage{graphicx}
\usepackage{epstopdf}
\usepackage{epsfig}
\usepackage{amssymb}
\usepackage[normalem]{ulem}
\usepackage{url}
\usepackage{float}
\usepackage{multirow}
\usepackage{colortbl}

\usepackage{microtype}

\aclfinalcopy 
\def\aclpaperid{xxx} 

\title{Inductively Representing Out-of-Knowledge-Graph Entities by \\ Optimal Estimation Under Translational Assumptions}
 
\author{
Damai Dai\textsuperscript{1},
Hua Zheng\textsuperscript{1},
Fuli Luo\textsuperscript{1},
Pengcheng Yang\textsuperscript{1},
\\
\textbf{Baobao Chang\textsuperscript{1,2},
Zhifang Sui\textsuperscript{1,2}}
\\
\textsuperscript{1}Key Lab of Computational Linguistics, School of EECS, Peking University 
\\
\textsuperscript{2}Peng Cheng Laboratory, China 
\\
\texttt{\{daidamai,zhenghua,luofuli,yang\_pc,chbb,szf\}@pku.edu.cn}
}

\date{}

\begin{document}
\maketitle
\begin{abstract}

Conventional Knowledge Graph Completion (KGC) assumes that all test entities appear during training. 
However, in real-world scenarios, Knowledge Graphs (KG) evolve fast with out-of-knowledge-graph (OOKG) entities added frequently, and we need to represent these entities efficiently. 
Most existing Knowledge Graph Embedding (KGE) methods cannot represent OOKG entities without costly retraining on the whole KG. 
To enhance efficiency, we propose a simple and effective method that inductively represents OOKG entities by their optimal estimation under translational assumptions. 
Given pretrained embeddings of the in-knowledge-graph (IKG) entities, our method needs no additional learning. 
Experimental results show that our method outperforms the state-of-the-art methods with higher efficiency on two KGC tasks with OOKG entities. 
\end{abstract}

\section{Introduction}

Knowledge Graphs (KG) play a pivotal role in various NLP tasks, but generally suffer from incompleteness. 
To address this problem, Knowledge Graph Completion (KGC) aims to predict missing relations in a KG based on Knowledge Graph Embeddings (KGE) of entities.  
Conventional KGE methods such as TransE~\citep{transe} and RotatE~\citep{rotate} achieve success in conventional KGC, which assumes that all test entities appear during training. 
However, in real-world scenarios, KGs evolve fast with out-of-knowledge-graph (OOKG) entities added frequently. 
To represent OOKG entities, most conventional KGE methods need to retrain on the whole KG frequently, which is extremely time-consuming. 
Faced with this problem, we are in urgent need of an efficient method to tackle KGC with OOKG entities. 

\begin{figure}[t]
	\centering
    \includegraphics[width=0.48\textwidth]{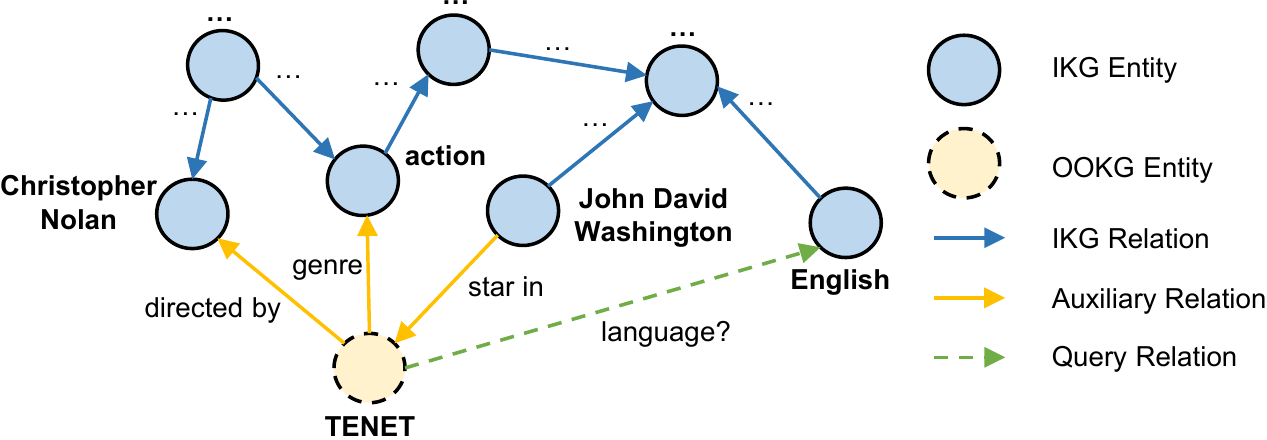}
	\caption{
	An example of KGC with OOKG entities. 
	When an OOKG entity ``\textbf{TENET}'' is added, we need to represent it efficiently via information of its IKG neighbors to predict its missing relations with other entities. 
	}
	\label{fig:problem}
\end{figure}

Figure~\ref{fig:problem} shows an example of KGC with OOKG entities. 
Based on an existing KG, a new movie ``\textbf{TENET}'' is added as an OOKG entity with some auxiliary relations that connect it with some in-knowledge-graph (IKG) entities. 
To predict the missing relations between ``\textbf{TENET}'' and other entities, we need to obtain its embedding first. 
Being aware that ``\textbf{TENET}'' is directed by ``\textbf{Christopher Nolan}'', is an ``\textbf{action}'' movie, and is starred by ``\textbf{John David Washington}'', we can combine these clues to profile ``\textbf{TENET}'' and estimate its embedding.
This embedding can then be used to predict whether its relation with ``\textbf{English}'' is ``language''. 

To represent OOKG entities via IKG neighbor information instead of retraining, \citet{gnn_ookb,lan,conv_layer,fcl_entity} adopt Graph Neural Networks (GNN) to aggregate IKG neighbors to obtain the OOKG entity embedding. 
Some other methods~\citep{dkrl,ikrl,conmask} utilize external resources such as entity descriptions or images instead of IKG neighbor information to avoid retraining. 
However, GNN models require relatively complex calculations, and high-quality external resources are hard and expensive to acquire. 

In this paper, we propose an inductive method that derives formulas to estimate OOKG entity embeddings from translational assumptions.  
Compared to existing methods, our method has simpler calculations and does not need external resources. 
For a triplet $(h, r, t)$, translational assumptions of KGE models suppose that embedding $\mathbf{h}$ can establish a connection with $\mathbf{t}$ via an $\mathbf{r}$-specific operation. 
Assuming that $h$ is OOKG and $t$ is IKG, we show that if a translational assumption can derive a specific formula to compute $\mathbf{h}$ via pretrained $\mathbf{t}$ and $\mathbf{r}$, then there will be no other candidate for $\mathbf{h}$ that better fits this translational assumption. 
Therefore, the computed $\mathbf{h}$ is the optimal estimation of the OOKG entity under this translational assumption. 
Among existing typical KGE models, we discover that translational assumptions of TransE and RotatE can derive specific estimation formulas. 
Therefore, based on them, we design two instances of our method called \textbf{InvTransE} and \textbf{InvRotatE}, respectively. 
Note that our estimation formulas are settled, so our method needs no additional learning when given pretrained IKG embeddings. 

Our contributions are summarized as follows: 
(1) 
We propose a simple and effective method to inductively represent OOKG entities by their optimal estimation under translational assumptions.  
(2) 
Our method needs no external resources. 
Given pretrained IKG embeddings, our method even needs no additional learning. 
(3) 
We evaluate our method on two KGC tasks with OOKG entities. 
Experimental results show that our method outperforms the state-of-the-art methods by a large margin with higher efficiency, and maintains a robust performance even under increasing OOKG entity ratios. 

\section{Methodology}

\subsection{Notations and Problem Formulation} \label{sec:notations}

Let $\mathcal{E}$ denote the IKG entity set and $\mathcal{R}$ denote the relation set. 
$\mathcal{K}_{\text{train}}$ is the training set where all entities are IKG. 
$\mathcal{K}_{\text{aux}}$ is the auxiliary set connecting OOKG and IKG entities when inferring, where each triplet contains an OOKG and an IKG entity. 
We define the $\mathcal{K}$-neighbor set of an entity $e$ as all its neighbor entities and relations in $\mathcal{K}$: $\mathcal{N}_{\mathcal{K}}(e)=\{(r, t) | (e, r, t) \in {\mathcal{K}}\} \cup \{(h, r) | (h, r, e) \in {\mathcal{K}}\}$. 

Using notations above, we formulate our problem as follows: Given $\mathcal{K}_{\text{aux}}$ and IKG embeddings pretrained on $\mathcal{K}_{\text{train}}$, we need to utilize them to represent an OOKG entity $e \not\in \mathcal{E}$ as an embedding. 
This embedding can then be used to tackle KGC with OOKG entities.

\subsection{Proposed Method}

\begin{figure}[t]
	\centering
    \includegraphics[width=0.475\textwidth]{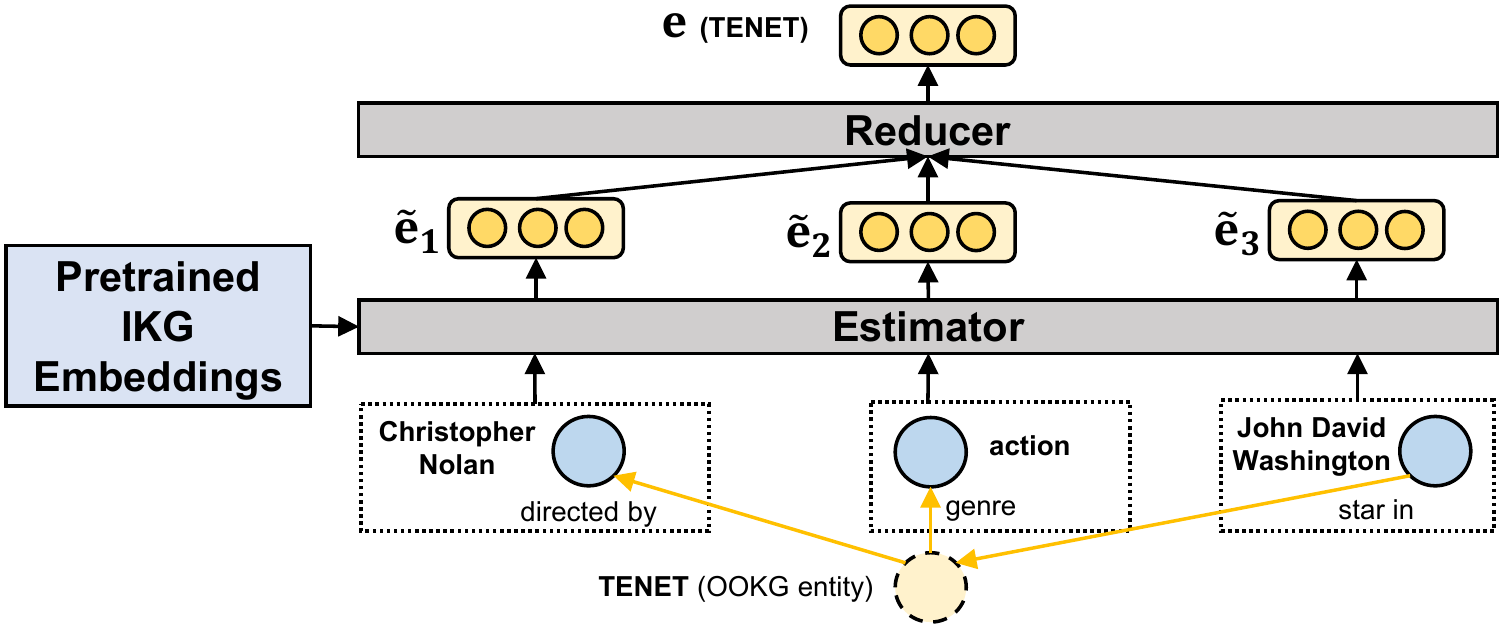}
	\caption{
	An illustration of our method. 
	It is composed of an estimator and a reducer. 
	}
	\label{fig:model}
\end{figure}

As shown in Figure~\ref{fig:model}, our proposed method is composed of an estimator and a reducer. 
The estimator aims to compute a set of candidate embeddings for an OOKG entity via its IKG neighbor information. 
The reducer aims to reduce these candidates to the final embedding of the OOKG entity. 

\subsubsection{Estimator}

For an OOKG entity $e$, given its IKG neighbors $\mathcal{N}_{\mathcal{K}_{\text{aux}}}(e)$ with pretrained embeddings, the estimator aims to compute a set of candidate embeddings. 
Except TransE and RotatE, other typical KGE models have relatively complex calculations in their translational assumptions. 
These complex calculations prevent their translational assumptions from deriving specific estimation formulas for OOKG entities.\footnote{Detailed proof is included in Appendix.} 
Therefore, we design two sets of estimation formulas based on TransE and RotatE, respectively.
To be specific, if $e$ is the head entity, we can obtain its optimal estimation $\widetilde{\mathbf{e}}$ by the following formulas:
\begin{equation*}
    \widetilde{\mathbf{e}} = \left\{
    \begin{aligned}
        & \mathbf{t} - \mathbf{r} , & \text{for } \textbf{InvTransE}, \\
        & \mathbf{t} \circ \mathbf{r}^{-1} , & \text{for } \textbf{InvRotatE},
    \end{aligned}
    \right.
\end{equation*}
where $\circ$ denotes the element-wise product, $\mathbf{r}^{-1}$ denotes the element-wise inversion. 

Otherwise, if $e$ is the tail entity, we can obtain its optimal estimation $\widetilde{\mathbf{e}}$ by the following formulas:
\begin{equation*}
    \widetilde{\mathbf{e}} = \left\{
    \begin{aligned}
        & \mathbf{h} + \mathbf{r} , & \text{for } \textbf{InvTransE}, \\
        & \mathbf{h} \circ \mathbf{r} , & \text{for } \textbf{InvRotatE}.
    \end{aligned}
    \right.
\end{equation*}

\subsubsection{Reducer} \label{sec:reducer}

After the estimator computes $\left|\mathcal{N}_{\mathcal{K}_{\text{aux}}}(e)\right|$ candidate embeddings, the reducer aims to reduce them to the final embedding of the OOKG entity by weighted average. 
We design two weighting functions. 

\textbf{Correlation-based weights} are query-aware. 
Inspired by \citet{lan}, we first use the conditional probability to model the correlation between two relations: 
\begin{equation*}
    P(r_{2} | r_{1}) = 
    \frac
    {
        \sum_{e \in \mathcal{E}}
        {
            \mathbbm{1} \left( r_{1}, r_{2} \in \mathcal{N}_{\mathcal{K}_{\text{train}}}\left(e\right) \right.)
        }
    }
    {
        \sum_{e \in \mathcal{E}}
        {
            \mathbbm{1} \left( r_{1} \in \mathcal{N}_{\mathcal{K}_{\text{train}}}\left(e\right) \right.)
        }
    }.
\end{equation*}

When the query relation $r_{q}$ is specified, we assign more weight to the candidate computed via a neighbor with a more relevant relation to $r_{q}$: 
\begin{equation*}
w_{\text{corr}}( \widetilde{\mathbf{e}} ) = 
\frac
{P \left( r_{\widetilde{\mathbf{e}}}|r_{q} \right) + P \left( r_{q}|r_{\widetilde{\mathbf{e}}} \right)}
{Z_{\text{corr}}}
, 
\end{equation*} 
where $Z_{\text{corr}}$ is the normalization factor, $r_{\widetilde{\mathbf{e}}}$ is the neighbor relation via which $\widetilde{\mathbf{e}}$ is computed.

\textbf{Degree-based weights} focus more on the entity with higher degree in the training set: 
\begin{equation*}
w_{\text{deg}}( \widetilde{\mathbf{e}} ) = 
\frac{\log{(d_{\widetilde{\mathbf{e}}} + \delta)}}
{Z_{\text{deg}}}
, 
\end{equation*}
where $Z_{\text{deg}}$ is the normalization factor, $d_{\widetilde{\mathbf{e}}}$ is the degree of the neighbor entity via which $\widetilde{\mathbf{e}}$ is computed, $\delta$ is a smoothing factor.

Based on these weighting functions, the final embedding of the OOKG entity $\mathbf{e}$ is computed by 
\begin{equation*}
\mathbf{e} = \sum_{\widetilde{\mathbf{e}} \in \mathcal{C}}~{\widetilde{\mathbf{e}} \cdot w_\text{corr/deg}( \widetilde{\mathbf{e}} )}
, 
\end{equation*}
where $\mathcal{C}$ denotes the candidate embedding set. 

\section{Experiments}

\subsection{Tasks and Datasets}

We conduct experiments on two KGC tasks with OOKG entities: link prediction and triplet classification. 
For link prediction, we use two datasets released by~\citet{lan} built based on FB15k~\citep{transe}. 
For triplet classification, we use nine datasets released by~\citet{gnn_ookb} built based on WN11~\citep{wn11}. 
All datasets are built for KGC with OOKG entities and composed of a training set, an auxiliary set, a validation set, and a test set. 
More details of these datasets are included in Appendix. 

\subsection{Experimental Settings}

We tune hyper-parameters for pretraining on the validation set. 
Generally, we use Adam~\citep{adam} with an initial learning rate of $10^{-3}$ as the optimizer and a batch size of $1,024$. 
For link prediction, we use an embedding dimension of $1,000$ and the correlation-based weights. 
For triplet classification, we use an embedding dimension of $300$ and the degree-based weights. 
Details of experimental settings are included in Appendix. 

\begin{table}[t]
\centering
\footnotesize
\setlength{\tabcolsep}{3pt}
\begin{tabular}{c | c c c | c c c }
\toprule
\multirow{2}{*}{\textbf{Method}} & \multicolumn{3}{c|}{\textbf{FB15k-Head-10}} & \multicolumn{3}{c}{\textbf{FB15k-Tail-10}} \\ 
 & MRR & H@10 & H@1 & MRR & H@10 & H@1 \\ 
\midrule
GNN-LSTM & 0.254 & 42.9 & 16.2 & 0.219 & 37.3 & 14.3 \\
GNN-MEAN & 0.310 & 48.0 & 22.2 & 0.251 & 41.0 & 17.1 \\
LAN & 0.394 & \uline{56.6} & 30.2 & 0.314 & 48.2 & 22.7 \\
\midrule
\textbf{InvTransE} & \textbf{0.462} & \textbf{60.4} & \textbf{38.5} & \uline{0.357} & \uline{48.7} & \uline{29.0} \\
\textbf{InvRotatE} & \uline{0.453} & \textbf{60.4} & \uline{36.9} & \textbf{0.362} & \textbf{49.1} & \textbf{29.3} \\
\bottomrule
\end{tabular}
\caption{
Evaluation results (MRR, Hits@k) of link prediction. 
\textbf{Bold} is the best. 
\uline{Underline} is the second best. 
}
\label{tab:lp_on_fb15k}
\end{table}

\begin{table}[t]
\centering
\footnotesize
\setlength{\tabcolsep}{2pt}
\begin{tabular}{c | c | c | c c c}
\toprule
\multirow{2}{*}{\textbf{Method}} & \multicolumn{1}{c|}{\textbf{WN11-Head}} & \multicolumn{1}{c|}{\textbf{WN11-Tail}} & \multicolumn{3}{c}{\textbf{WN11-Both}} \\ 
 & 3000 & 3000 & 1000 & 3000 & 5000 \\ 
\midrule
ConvLayer & - & - & 74.9 & - & 64.6 \\
FCLEntity & 82.6 & 72.1 & - & 68.6 & - \\
GNN-LSTM & 83.5 & 71.4 & 78.5 & 71.6 & 65.8 \\
GNN-MEAN & 84.3 & 75.2 & 83.0 & 73.3 & 68.2 \\
LAN & 85.2 & \uline{78.8} & 83.3 & \uline{76.9} & \uline{70.6} \\
\midrule
\textbf{InvTransE} & \textbf{87.8} & \textbf{80.1} & \textbf{86.3} & \textbf{78.4} & \textbf{74.6} \\
\textbf{InvRotatE} & \uline{86.9} & \textbf{80.1} & \uline{84.2} & 75.0 & \uline{70.6} \\
\bottomrule
\end{tabular}
\caption{
Evaluation results (Accuracy) of triplet classification. 
\textbf{Bold} is the best. 
\uline{Underline} is the second best.
}
\label{tab:tc_on_wn11}
\end{table}

\subsection{Baselines}

For link prediction, we compare our method with three GNN-based baselines. 
\textbf{GNN-MEAN}~\cite{gnn_ookb} uses a mean function to aggregate neighbors. 
\textbf{GNN-LSTM} adopts LSTM for aggregation. 
\textbf{LAN}~\cite{lan} adopts a both rule- and network-based attention mechanism for aggregation and maintains the best performance so far. 
For triplet classification, we compare with two more GNN-based baselines. 
\textbf{ConvLayer}~\cite{conv_layer} uses convolutional layers as the transition function. 
\textbf{FCLEntity}~\cite{fcl_entity} uses fully-connected networks as the transition function and adopts an attention-based aggregation.  

\subsection{Evaluation Metrics}

For link prediction, we use Mean Reciprocal Rank (MRR) and the proportion of ground truth entities ranked in top-k (Hits@k, $k \in \left\{ 1, 10 \right\}$). 
All the metrics are filtered versions that exclude false negative candidates. 
For triplet classification, we use Accuracy. 
We determine relation-specific thresholds $\delta_{r}$ by maximizing the accuracy on the validation set. 

\subsection{Main Results}

Evaluation results of link prediction are shown in Table~\ref{tab:lp_on_fb15k}. 
From the table, we have the following observations: 
(1) Both instances of our method significantly outperform all baselines since our estimation formulas are optimal under translational assumptions. 
(2) GNN-LSTM performs the worst since neighbors are unordered but LSTM captures ordered information. 
(3) LAN is the best baseline since it adopts a complex attention mechanism to aggregate neighbors more comprehensively. 
For triplet classification, due to space limitation, we show the main part of the results in Table~\ref{tab:tc_on_wn11} and the complete results in Appendix. 
From Table~\ref{tab:tc_on_wn11}, we find that our method outperforms all baselines on all datasets due to our optimal estimation. 

\begin{table}[t]
\centering
\footnotesize
\setlength{\tabcolsep}{8pt}
\begin{tabular}{c | c c c }
\toprule
\textbf{Method} & MRR & H@10 & H@1 \\ 
\midrule
InvTransE (Full) & 0.462 & 60.4 & 38.5 \\
\midrule
Up to 32 Neighbors & 0.447 & 59.2 & 37.2 \\
Up to 8 Neighbors & 0.386 & 52.0 & 31.3 \\
Only 1 Neighbor & 0.246 & 37.9 & 18.1 \\
\midrule
Uniform Weights & 0.361 & 52.0 & 28.1 \\
\bottomrule
\end{tabular}
\caption{Ablation experiment results for InvTransE on the FB15k-Head-10 dataset of link prediction. }
\label{tab:ablation}
\end{table}

\subsection{Analysis} 

\noindent\textbf{How does the number of neighbors impact the performance?}
We randomly select up to $k \in \left\{1, 8, 32\right\}$ IKG neighbors of OOKG entities to use. 
As shown in Table~\ref{tab:ablation}, as the number of used neighbors decreases, the performance drops. 
This suggests that using more neighbors can enhance the robustness and thus lead to better performance. 

\noindent\textbf{Do our weighting functions matter?}
We attempt to reduce candidates with uniform weights. 
As shown in Table~\ref{tab:ablation}, the performance without our weighting functions drops dramatically. 
This verifies the effectiveness of our weighting functions. 

\begin{figure}[t]
	\centering
    \includegraphics[width=0.48\textwidth]{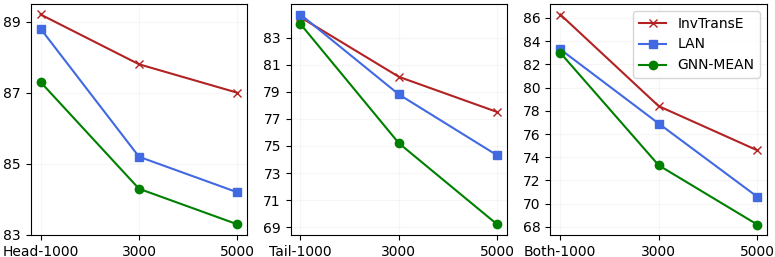}
	\caption{
	Results under increasing OOKG entity ratios.  
	}
	\label{fig:ratio}
\end{figure}

\noindent\textbf{How does our method perform under increasing OOKG entity ratios?} 
We compare the triplet classification results of InvTransE, LAN, and GNN-MEAN under increasing OOKG entity ratios in Figure~\ref{fig:ratio}. 
We find that, as the OOKG entity ratio increases, the performance of our method drops the slowest. 
This suggests that our method is more robust to increasing OOKG entity ratios. 

\noindent\textbf{Is our method more efficient?} 
We compare InvTransE with LAN to highlight the efficiency of our method.  
Theoretically, LAN requires $\mathcal{O}(md^2)$ to represent an entity, where $m$ is the number of neighbors and $d$ is the embedding dimension. 
By contrast, InvTransE requires only $\mathcal{O}(d)$ and $\mathcal{O}(md)$ to represent an IKG and an OOKG entity, respectively. 
Empirically, under similar configurations, LAN costs about 15 times the time of InvTransE to train a model for triplet classification. 
This verifies that our simple method is much more efficient.  

\section{Related Work}

Conventional transductive KGE methods map entities and relations to embeddings, and then use score functions to measure the salience of triplets. 
TransE~\citep{transe} pioneers translational distance methods and is the most widely-used one. 
It derives a series of translational distance methods, such as TransH~\citep{transh}, TransR~\citep{transr}, and RotatE~\citep{rotate}. 
Besides, semantic matching methods form another mainstream~\citep{rescal, distmult, complex, hole, tucker}. 
These transductive KGE methods achieve success in conventional KGC, but fail to directly represent OOKG entities efficiently. 

To represent OOKG entities more efficiently, some inductive methods adopt GNN to aggregate IKG neighbors to inductively produce embeddings for OOKG entities~\citep{gnn_ookb,lan,conv_layer,fcl_entity}. 
These methods are effective, but need relatively complex calculations. 
Other inductive methods incorporate external resources to enrich embeddings and represent OOKG entities via only external resources~\citep{dkrl, conmask, ikrl}. 
However, high-quality external resources are hard and expensive to acquire. 

\section{Conclusion}
 
This paper aims to address the problem of efficiently representing OOKG entities. 
We propose a simple and effective method that inductively represents OOKG entities by their optimal estimation under translational assumptions. 
Given pretrained IKG embeddings, our method needs no additional learning. 
Experimental results on two KGC tasks with OOKG entities show that our method outperforms the state-of-the-art methods by a large margin with higher efficiency, and maintains a robust performance under increasing OOKG entity ratios. 

\bibliography{anthology,emnlp2020}
\bibliographystyle{acl_natbib}

\onecolumn
\appendix
\appendixpage

\section{Which Translational Assumptions Can Derive Specific Estimation Formulas for OOKG entities?} 

For a triplet $(h, r, t)$, translational assumptions of KGE models suppose that $\mathbf{h}$ can establish a connection with $\mathbf{t}$ via an $\mathbf{r}$-specific operation, which can be formulated by the following equation: 
\begin{equation}\label{eq:translational_assumptions}
    \mathcal{F}_\mathbf{r} (\mathbf{h}, \mathbf{t}) = 0, 
\end{equation}
where $\mathcal{F}_\mathbf{r} (\cdot)$ is an $\mathbf{r}$-specific function that is determined by the specific KGE model. 
Without loss of generality, we may assume that $h$ is an OOKG entity and $t$ is an IKG entity. 
Under a translational assumption, we can obtain a specific estimation formula for $\mathbf{h}$ if and only if (1) we regard $\mathbf{h}$ as unknown, and its solution in Equation~\ref{eq:translational_assumptions} exists, (2) the solution is unique. 
If the above two conditions hold, the unique solution of $\mathbf{h}$ is the optimal estimation under the translational assumption, since no other candidate for $\mathbf{h}$ can better fit Equation~\ref{eq:translational_assumptions}. 
In the following parts, we analyze translational assumptions of four KGE models (TransE, RotatE, TransH, TransR) as examples. 

\subsection{TransE}
For TransE, its translational assumption is formulated by
\begin{equation}
    \mathcal{F}_\mathbf{r} (\mathbf{h}, \mathbf{t}) = \left\| \mathbf{h}+\mathbf{r}-\mathbf{t} \right\|_{1/2} = 0. 
\end{equation}
In this case, we can obtain a unique solution of $\mathbf{h}$ by the following steps: 
\begin{align}
    & \left\| \mathbf{h} + \mathbf{r} - \mathbf{t} \right\|_{1/2} = 0, 
    \\
    & \Longrightarrow \mathbf{h} + \mathbf{r} - \mathbf{t} = \mathbf{0}, 
    \\
    & \Longrightarrow \mathbf{h} = \mathbf{t} - \mathbf{r}.
\end{align}
This computed $\mathbf{h}$ is the optimal estimation under the translational assumption. 

\subsection{RotatE}
For RotatE, its translational assumption is formulated by
\begin{equation}
    \mathcal{F}_\mathbf{r} (\mathbf{h}, \mathbf{t}) = \left\| \mathbf{h} \circ \mathbf{r}-\mathbf{t} \right\|_{1/2} = 0. 
\end{equation}
In this case, we can obtain a unique solution of $\mathbf{h}$ by the following steps: 
\begin{align}
    & \left\| \mathbf{h} \circ \mathbf{r} - \mathbf{t} \right\|_{1/2} = 0, 
    \\
    & \Longrightarrow \mathbf{h} \circ \mathbf{r} - \mathbf{t} = \mathbf{0}, 
    \\
    & \Longrightarrow \mathbf{h} = \mathbf{t} \circ \mathbf{r}^{-1}.
\end{align}
This computed $\mathbf{h}$ is the optimal estimation under the translational assumption. 

\subsection{TransH}
For TransH, its translational assumption is formulated by
\begin{equation}
    \mathcal{F}_\mathbf{r} (\mathbf{h}, \mathbf{t}) = \left\| (\mathbf{h}-\mathbf{w}_{r}^{\top} \mathbf{h} \mathbf{w}_{r})+\mathbf{r}-(\mathbf{t}-\mathbf{w}_{r}^{\top} \mathbf{t} \mathbf{w}_{r}) \right\|_{1/2} = 0, 
\end{equation}
where $\mathbf{w}_{r}$ is the unit normal vector of the plane $P$ that $\mathbf{r}$ lies on. 
From the translational assumption, we can derive the following equations: 
\begin{align}
    & \left\| (\mathbf{h}-\mathbf{w}_{r}^{\top} \mathbf{h} \mathbf{w}_{r})+\mathbf{r}-(\mathbf{t}-\mathbf{w}_{r}^{\top} \mathbf{t} \mathbf{w}_{r}) \right\|_{1/2} = 0,
    \\
    & \Longrightarrow (\mathbf{h}-\mathbf{w}_{r}^{\top} \mathbf{h} \mathbf{w}_{r})+\mathbf{r}-(\mathbf{t}-\mathbf{w}_{r}^{\top} \mathbf{t} \mathbf{w}_{r}) = \mathbf{0}, 
    \\
    & \Longrightarrow (\mathbf{h}-\mathbf{w}_{r}^{\top} \mathbf{h} \mathbf{w}_{r}) = (\mathbf{t}-\mathbf{w}_{r}^{\top} \mathbf{t} \mathbf{w}_{r}) - \mathbf{r} \triangleq \mathbf{v}. 
\end{align}
$\mathbf{h}-\mathbf{w}_{r}^{\top} \mathbf{h} \mathbf{w}_{r}$ is the projection of $\mathbf{h}$ on the plane $P$. 
From the translational assumption, we can only deduce that the projection of $\mathbf{h}$ is equal to $\mathbf{v}$. 
However, there exist infinitely many possible $\mathbf{h}$ that can satisfy this condition. 
Therefore, the solution of $\mathbf{h}$ is not unique, and we cannot obtain a specific estimation formula from the translational assumption of TransH.

\subsection{TransR}
For TransR, its translational assumption is formulated by
\begin{equation}
    \mathcal{F}_\mathbf{r} (\mathbf{h}, \mathbf{t}) = \left\| \mathbf{M}_{r}\mathbf{h} + \mathbf{r} - \mathbf{M}_{r}\mathbf{t} \right\|_{1/2} = 0, 
\end{equation}
where $\mathbf{M}_{r}$ is an r-specific matrix. 
From the translational assumption, we can derive the following equations: 
\begin{align}
    &  \left\| \mathbf{M}_{r}\mathbf{h} + \mathbf{r} - \mathbf{M}_{r}\mathbf{t} \right\|_{1/2} = 0,
    \\
    & \Longrightarrow \mathbf{M}_{r}\mathbf{h} + \mathbf{r} - \mathbf{M}_{r}\mathbf{t} = \mathbf{0}, 
    \\
    & \Longrightarrow \mathbf{M}_{r}\mathbf{h} = \mathbf{M}_{r}\mathbf{t} - \mathbf{r} \triangleq \mathbf{v}. 
\end{align}
In this case, we derive a system of linear equations from the translational assumption. 
In this system, there exists a unique solution for $\mathbf{h}$ if and only if the rank of the coefficient matrix $\mathbf{M}_{r}$ is equal to the rank of the augmented matrix $\left[ \mathbf{M}_{r}; \mathbf{v} \right]$. 
However, $\mathbf{M}_{r}$ is automatically learned by TransR without this restriction. 
Therefore, we cannot guarantee that there exists a unique solution for $\mathbf{h}$, and we cannot obtain a specific estimation formula from the translational assumption of TransR. 

\section{Details of Datasets} 

\begin{table*}[ht]
\centering
\setlength{\tabcolsep}{4pt}
\begin{tabular}{l | c c c c | c c c}
\toprule
Dataset & $\left| \mathcal{K}_{\text{train}} \right|$ & $\left| \mathcal{K}_{\text{valid}} \right|$ & $\left| \mathcal{K}_{\text{aux}} \right|$ & $\left| \mathcal{K}_{\text{test}} \right|$ & $\left| \mathcal{R} \right|$ & $\left| \mathcal{E} \right|$ & $\left| \mathcal{E}^{\prime} \right|$ \\
\midrule
\midrule
FB15k-Head-10 & 108,854 & 11,339 & 249,798 & 2,811 & 1,170 & 10,336 & 2,082 \\
FB15k-Tail-10 & ~~99,783 & 10,190 & 261,341 & 2,987 & 1,126 & 10,603 & 1,934 \\
\midrule
\midrule
WN11-Head-1000 & 108,197 & ~~4,561 & ~~1,938 & ~~~955 & 11 & 37,700 & ~~~340 \\
WN11-Head-3000 & ~~99,963 & ~~4,068 & ~~5,311 & 2,686 & 11 & 36,646 & ~~~985 \\
WN11-Head-5000 & ~~92,309 & ~~3,688 & ~~8,048 & 4,252 & 11 & 35,560 & 1,638 \\
\midrule
WN11-Tail-1000 & ~~96,968 & ~~3,864 & ~~6,674 & ~~~852 & 11 & 36,771 & ~~~811 \\
WN11-Tail-3000 & ~~78,812 & ~~2,851 & 12,824 & 2,061 & 11 & 33,800 & 1,874 \\
WN11-Tail-5000 & ~~68,040 & ~~2,258 & 15,414 & 2,968 & 11 & 31,311 & 2,589 \\
\midrule
WN11-Both-1000 & ~~93,683 & ~~3,625 & ~~7,875 & ~~~873 & 11 & 36,277 & 1,136 \\
WN11-Both-3000 & ~~71,618 & ~~2,436 & 14,453 & 2,242 & 11 & 32,254 & 2,805 \\
WN11-Both-5000 & ~~58,923 & ~~1,788 & 16,660 & 3,218 & 11 & 28,979 & 3,934 \\
\bottomrule
\end{tabular}
\caption{
Statistics of datasets with OOKG entities. 
These datasets are built based on FB15k or WN11 and named in the form of ``Base-Pos-Num''. 
Base denotes the based datasets. 
Pos denotes the position of OOKG entities in test triplets. 
Num distinguishes different numbers of OOKG entities represented by $\left| \mathcal{E}^{\prime} \right|$. 
}
\label{tab:datasets}
\end{table*}

For link prediction, we use two datasets released by~\citet{lan}: FB15k-Head-10 and FB15k-Tail-10. 
These two datasets are built based on FB15k~\citep{transe}. 
For triplet classification, we use nine datasets released by~\citet{gnn_ookb}: WN11-Head-1000, WN11-Head-3000, WN11-Head-5000, WN11-Tail-1000, WN11-Tail-3000, WN11-Tail-5000, WN11-Both-1000, WN11-Both-3000, and WN11-Both-5000. 
These nine datasets are built based on WN11~\citep{wn11}. 
Each of the datasets mentioned above is composed of four sets: a training set, an auxiliary set, a validation set, and a test set. 
Each triplet in the training and validation sets contains only IKG entities. 
Each triplet in the auxiliary set contains an OOKG entity and an IKG entity. 
Each triplet in the test set contains at least one OOKG entity. 
The statistics of the datasets are shown in Table~\ref{tab:datasets}. 

\section{Details of Experimental Settings}

\begin{table*}[ht]
\centering
\setlength{\tabcolsep}{7pt}
\begin{tabular}{l | c c c c c c}
\toprule
Datasets & $d$ & $\gamma$ & $\alpha$ & $n$ & L2 & Training Steps \\
\midrule
FB15k-based & 1,000 & 24.0 & 1.0 & 256 & N/A & 100,000 \\
WN11-based & 300 & 0.5 & 1.0 & 128 & $10^{-5}$ & 20,000 \\
\bottomrule
\end{tabular}
\caption{
Hyper-parameters for two categories of datasets. 
We use the same hyper-parameters for two FB15k-based datasets and the same hyper-parameters for nine WN11-based datasets. 
On each dataset, we use the same hyper-parameters for two pretrained models. 
$d$ denotes the embedding dimension. 
$\gamma$ denotes the margin. 
$\alpha$ denotes the sampling temperature. 
$n$ denotes the negative sampling size. 
L2 denotes the parameter of L2 regularization, where N/A means no regularization. }
\label{tab:paras}
\end{table*}

To pretrain the TransE and RotatE models, we adopt the self-adversarial negative sampling loss proposed by~\citet{rotate} in consideration of its good performance on training TransE and RotatE. 
The self-adversarial negative sampling loss $L$ is formulated as
\begin{equation*}
L=
-\log \sigma \left( \gamma - \mathcal{D}\left(\mathbf{h}, \mathbf{r}, \mathbf{t}\right) \right)
-\sum_{i=1}^{n}~p\left( h_{i}^{\prime}, r, t_{i}^{\prime} \right) \log \sigma\left(\mathcal{D}\left(\mathbf{h}_{i}^{\prime}, \mathbf{r}, \mathbf{t}_{i}^{\prime} \right)-\gamma \right)
, 
\end{equation*}
where $\sigma$ is the sigmoid function, $\gamma$ is the margin, $n$ is the negative sampling size and $\left( h_{i}^{\prime}, r, t_{i}^{\prime} \right)$ is the i-th negative sample triplet. 
$\mathcal{D}\left( \cdot  \right)$ is the distance function. 
$\mathcal{D}\left( \mathbf{h}, \mathbf{r}, \mathbf{t}  \right)$ is equal to $\left\| \mathbf{h}+\mathbf{r}-\mathbf{t} \right\|_{1/2}$ for TransE and is equal to $\left\| \mathbf{h} \circ \mathbf{r}-\mathbf{t} \right\|_{1/2}$ for RotatE. 
$p$ is the self-adversarial weight function which gives more weight to the high-scored negative samples:  
\begin{equation*}
    p\left( h_{i}^{\prime}, r, t_{i}^{\prime} \right) \propto \exp \left( \alpha \cdot \mathcal{F}\left(\mathbf{h}_{i}^{\prime}, \mathbf{r}, \mathbf{t}_{i}^{\prime}\right)\right),
\end{equation*}
where $\alpha$ is a hyper-parameter called sampling temperature to be tuned. 
$\mathcal{F}(\cdot)$ is the score function that is equal to $-\mathcal{D}(\cdot)$. 

We conduct each experiment on a single Nvidia Geforce GTX-1080Ti GPU and tune hyper-parameters on the validation set. 
Generally, we set the batch size to $1024$ and use Adam~\citep{adam} with an initial learning rate of $10^{-3}$ as the optimizer. 
We choose the correlation-based weights for link prediction and choose the degree-based weights with a smoothing factor of $0.1$ for triplet classification. 
Other hyper-parameters are shown in Table \ref{tab:paras}. 

\section{Complete Evaluation Results of Triplet Classification}

\begin{table*}[ht]
\centering
\footnotesize
\setlength{\tabcolsep}{6pt}
\begin{tabular}{c | c c c | c c c | c c c}
\toprule
\multirow{2}{*}{\textbf{Method}} & \multicolumn{3}{c|}{\textbf{WN11-Head}} & \multicolumn{3}{c|}{\textbf{WN11-Tail}} & \multicolumn{3}{c}{\textbf{WN11-Both}} \\ 
 & 1000 & 3000 & 5000 & 1000 & 3000 & 5000 & 1000 & 3000 & 5000 \\ 
\midrule
ConvLayer & - & - & - & - & - & - & 74.9 & - & 64.6 \\
FCLEntity & - & 82.6 & - & - & 72.1 & - & - & 68.6 & - \\
GNN-LSTM & 87.0 & 83.5 & 81.8 & 82.9 & 71.4 & 63.1 & 78.5 & 71.6 & 65.8 \\
GNN-MEAN & 87.3 & 84.3 & 83.3 & 84.0 & 75.2 & 69.2 & 83.0 & 73.3 & 68.2 \\
LAN & \uline{88.8} & 85.2 & 84.2 & \textbf{84.7} & \uline{78.8} & 74.3 & 83.3 & \uline{76.9} & \uline{70.6} \\
\midrule
\textbf{InvTransE} & \textbf{89.2} & \textbf{87.8} & \textbf{87.0} & \uline{84.5} & \textbf{80.1} & \textbf{77.5} & \textbf{86.3} & \textbf{78.4} & \textbf{74.6} \\
\textbf{InvRotatE} & 88.6 & \uline{86.9} & \uline{86.5} & \textbf{84.7} & \textbf{80.1} & \uline{75.8} & \uline{84.2} & 75.0 & \uline{70.6} \\
\bottomrule
\end{tabular}
\caption{
Complete evaluation results (Accuracy) of triplet classification. 
\textbf{Bold} is the best. \uline{Underline} is the second best. 
The results of all five baselines are taken from their original papers. 
}
\label{tab:complete_rlts_tc}
\end{table*}

\end{document}